%% file: main.tex
\documentclass[runningheads]{llncs}

\usepackage{eccv}

\usepackage{eccvabbrv}
\usepackage{wrapfig}
\usepackage{xcolor}
\usepackage{multirow}
\usepackage{colortbl}
\usepackage{pifont}
\usepackage{graphicx}
\usepackage{booktabs}
\usepackage{amsmath}
\usepackage{color}
\usepackage{makecell}
\usepackage{adjustbox}
\usepackage[accsupp]{axessibility}  

\definecolor{myblue}{HTML}{C2D7F4}
\definecolor{mygray}{HTML}{eeeaeb}
\definecolor{mypurple}{HTML}{e8d9e9}
\definecolor{lightblue}{rgb}{0.9, 0.95, 1.0}
\definecolor{highlight}{rgb}{0.8, 0.9, 1.0}
\newcommand{\cmark}{\ding{51}}
\newcommand{\xmark}{\ding{55}}

\usepackage[pagebackref,breaklinks,colorlinks,citecolor=eccvblue]{hyperref}
\usepackage{hyperref}

\usepackage{orcidlink}

\begin{document}

\title{\Large DriveFine: Refining-Augmented Masked Diffusion VLA for Precise and Robust Driving} 

\titlerunning{DriveFine: Refining-Augmented ...}

\author{
Chenxu Dang\inst{1,2,3}\thanks{Completed during the internship at Xiaomi EV and AIR.  \textsuperscript{$\dagger$} Corresponding authors.} \and
Sining Ang\inst{3} \and
Yongkang Li\inst{1} \and
Haochen Tian\inst{2} \and
Jie Wang\inst{2} \and
Guang Li\inst{2} \and
Hangjun Ye\inst{2} \and
Jie Ma\inst{1} \and
Long Chen\inst{2}\textsuperscript{$^\dagger$} \and
Yan Wang\inst{3}\textsuperscript{$^\dagger$}
}

\authorrunning{C. Dang et al.}

\institute{Huazhong University of Science and Technology \and
Xiaomi EV\\
\and
Institute for AI Industry Research (AIR), Tsinghua University\\
}

\maketitle

\begin{abstract}
Vision-Language-Action (VLA) models for autonomous driving increasingly adopt generative planners trained with imitation learning followed by reinforcement learning. Diffusion-based planners suffer from modality alignment difficulties, low training efficiency, and limited generalization. Token-based planners are plagued by cumulative causal errors and irreversible decoding. In summary, the two dominant paradigms exhibit complementary strengths and weaknesses.
In this paper, we propose \textbf{DriveFine}, a masked diffusion VLA model that combines flexible decoding with self-correction capabilities. In particular, we design a novel plug-and-play block-MoE, which seamlessly injects a refinement expert on top of the generation expert. By enabling explicit expert selection during inference and gradient blocking during training, the two experts are fully decoupled, preserving the foundational capabilities and generic patterns of the pretrained weights, which highlights the flexibility and extensibility of the block-MoE design. Furthermore, we design a hybrid reinforcement learning strategy that encourages effective exploration of refinement expert while maintaining training stability. Extensive experiments on NAVSIM v1, v2, and Navhard benchmarks demonstrate that DriveFine exhibits strong efficacy and robustness. The code will be released at \url{https://github.com/MSunDYY/DriveFine}.
\end{abstract}

\input{secs/introduction}
\input{secs/related_work}

\input{secs/method}
\input{secs/experiments}

\section{Conclusion}
In this paper, we conduct a comprehensive analysis of the two dominant VLA planners for autonomous driving: diffusion-based and token-based paradigms, highlighting their respective strengths and limitations. We further explore leveraging masked diffusion LLMs as a potential solution to mitigate their shortcomings. Building on these insights, we propose DriveFine, which features a plug-and-play block-MoE architecture combined with a hybrid reinforcement training strategy to inject refinement capabilities into token-based VLAs. We evaluate DriveFine on NAVSIMv1, NAVSIMv2, and the more challenging Navhard benchmarks, and demonstrate its effectiveness and robustness through extensive ablation studies and comparative analyses. We hope our findings and contributions will provide valuable insights for the community.



%
%
\bibliographystyle{splncs04}
\bibliography{main}
\end{document}

%% file: secs/introduction.tex
\section{Introduction}
Vision-Language-Action (VLA) systems for autonomous driving (AD) integrate sensor observations and textual instructions, with a planner responsible for generating driving actions or trajectories. Early deterministic planners (e.g., MLP-based\cite{uniad,vad,sparsead} or anchor-based classifiers\cite{vadv2,hydra-mdp}) tended to imitate a single expert trajectory, which struggle with distribution shift and cannot capture the inherently multi-modal nature of driving.


Recently, non-deterministic generative planners have emerged as the dominant paradigms in AD VLAs.
They predict actions as probability distributions, effectively capturing the multimodal nature of driving behavior.
Moreover, their inherent sampling capability encourages active exploration and enables seamless integration with rule-driven reinforcement learning strategies, such as GRPO, to guide policy learning. Current state-of-the-art generative VLAs can be categorized into diffusion-based models\cite{recogdrive,sgdrive,diffusiondrive,diffusiondrivev2} with continuous action modeling and token-based planners\cite{opendrivevla,autovla,adathinkdrive} with discrete action representations.

\begin{figure}[t]
    \centering
    \includegraphics[width=1\linewidth]{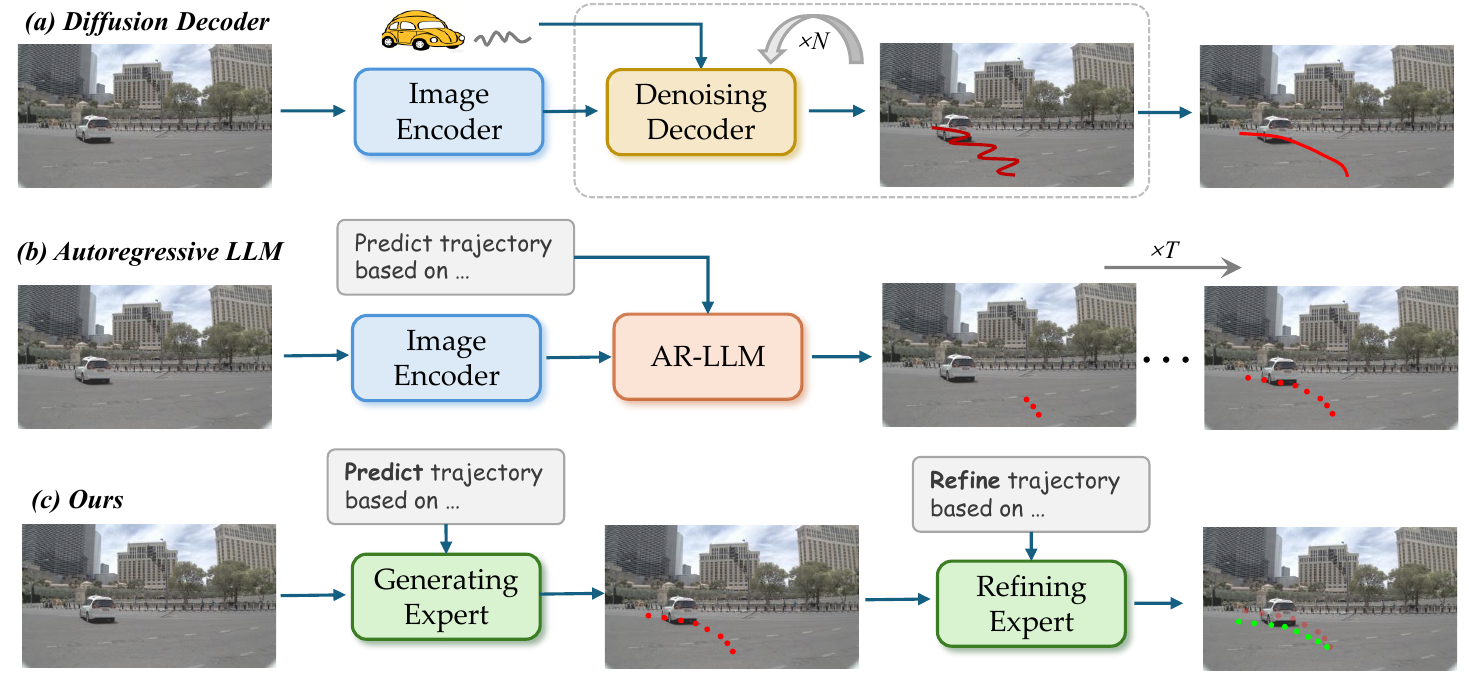}
    \caption{Comparison of decoding mechanisms for Action Tokens in Generative VLA Models. (a) Parallel refinement with multi-step denoising. (b) Token-by-token decoding. (c) Generate first in parallel, then refine.}
    \label{fig:comparison}
\end{figure}

\begin{wrapfigure}{r}{0.45\textwidth}
    \vspace{-10pt} 
    \centering
    \includegraphics[width=\linewidth]{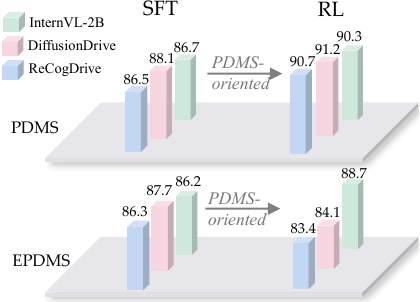}
    \caption{PDMS-oriented RFT.}
    \label{fig:reward-haking}
    \vspace{-10pt}
\end{wrapfigure}
(1) \textbf{Diffusion-based VLAs}, as shown in Fig.~\ref{fig:comparison}(a) construct a Markov chain and iteratively refine noisy trajectories by predicting their mean and variance.




Despite the efficiency enabled by parallel decoding, the additional diffusion Transformer hinders cross-modal alignment, leading to inefficient training that typically requires hundreds of epochs.

Moreover, diffusion-based planners are inherently conditional generators, which limits their robustness and generalization ability,
as evidenced empirically in Fig. \ref{fig:reward-haking}: when optimized with PDMS-oriented reinforcement fine-tuning, diffusion-based planners\cite{recogdrive,diffusiondrive} suffer a significant drop in EPDMS. Here, both PDMS and EPDMS are metrics provided by NAVSIM\cite{navsim}, see \ref{sec: navsim} for details. We attribute this degradation to the weak coupling between diffusion planners and the VLM, which induces reward hacking and the loss of pretrained knowledge, substantially limiting their practical applicability.


(2) \textbf{Token-based VLAs}, as illustrated in Fig. \ref{fig:comparison}(b), decode actions into tokens autoregressively within a predefined vocabulary, achieving a unified representation across vision, language, and action. As shown in Fig. \ref{fig:reward-haking}, the PDMS-oriented RFT for InternVL\cite{internvl} leads to simultaneous improvements in both PDMS and EPDMS, exhibiting its stronger generalization and extensibility.
\begin{figure}[t]
    \centering
    \includegraphics[width=1.0\linewidth]{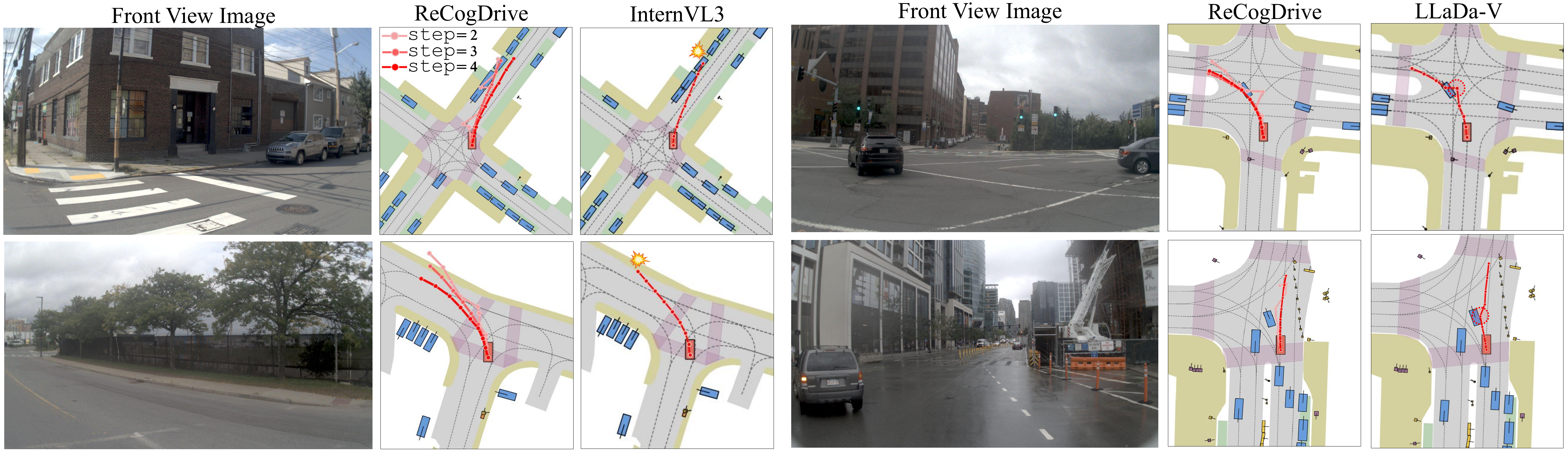}
    \caption{Failure cases caused by irreversible decoding in token-based VLAs.}
    \label{fig:vis_com}
\end{figure}






However, token-based VLAs\cite{adathinkdrive,autovla} generally lag behind diffusion-based counterparts\cite{recogdrive,diffusiondrivev2} in both performance and efficiency. This is mainly due to their causal attention and fixed token-by-token decoding, which is computationally costly and prone to error accumulation during inference. More critically, they inherit the irreversible decoding property of LLMs: decoded tokens cannot be modified once committed. Planning, however, is highly sensitive to noise: even point-level deviations can cause the entire trajectory failure, such as collision or off-road driving (Fig. \ref{fig:vis_com}).

Very recent works\cite{reflectdrive,wam-diff} explored masked diffusion LLMs\cite{llada,lavida} (dLLMs) featuring more flexible decoding orders for driving. However, this flexibility aggravates the irreversible decoding problems: as shown in Fig. \ref{fig:vis_com}, tokens decoded early lack global consistency constraints, are more likely to become outliers, and cannot be revised afterward, leading to trajectory-level failure. In contrast, diffusion-based planners iteratively refine the trajectory, enabling successive refinement and thereby ensuring high-quality trajectory generation.

Clearly, both VLA planners exhibit complementary strengths and weaknesses, motivating the exploration of a model harnessing the advantages of both. 

In this paper, we propose \textbf{DriveFine}, pioneering the explicit injection of token-VLA’s refining capability for more precise and robust driving. We adopt a pretrained multi-modal masked Diffusion LLM (LaViDa\cite{lavida} with LLaDA\cite{llada} as LLM) as our base planner, considering its several benefits compared with AR LLMs: parallel decoding for efficiency, bidirectional attention for richer context modeling, and a flexible decoding strategy that facilitates adaptive learning.

The refining of token-VLAs is far from trivial and must adhere to several principles: preserving the base VLM’s original training and inference paradigms to prevent collapse of foundational capabilities; minimizing extra overhead in computation and parameters; remaining decoupled from trajectory generation to avoid interference. Evidently, this presents substantial challenges. 

To this end, we design a block-wise Mixture-of-Experts (MoE) architecture. Specifically, the majority of LLaDA blocks serve as shared experts for common contexts, while the remaining blocks are explicitly partitioned into generation experts and refinement experts. During inference, task-specific experts are proactively selected. During training, gradient flow from the refining branch is strictly confined to the refinement experts, decoupling from the generation experts. This explicit isolation preserves the foundational capabilities of the generation experts, effectively preventing mode collapse and cross-task interference.

To align with the dominant pipeline, we further introduce an online-offline hybrid reinforcement learning paradigm. The generation expert samples a group of trajectories, which are reinforced via GRPO. Simultaneously, these trajectories are paired into offline anchor-target trajectory tuples. In parallel, the refinement expert actively refines the above trajectories, computing the associated rewards and generating online trajectory pairs, which, together with offline pairs, jointly supervise the refinement expert. Experimental results demonstrate that the block-wise MOE significantly enhances trajectory quality with limited parameter increase and slight inference overhead, raising the performance ceiling of token-based VLAs. We summarize our core contributions as follows:
\begin{itemize}
    \item We provide a thorough analysis of the strengths and weaknesses of mainstream diffusion-based and token-based VLAs.
    \item We propose DriveFine, featuring a plug-and-play block-wise Mixture-of-Experts (block-MoE) that injects refining capability into token-based VLAs at minimal cost.
    \item We design a targeted hybrid reinforcement learning strategy to further elevate DriveFine’s performance ceiling.
    \item Extensive experiments demonstrate that DriveFine consistently achieves state-of-the-art performance on NavSim v1, v2, and Navhard benchmarks.
\end{itemize}

%% file: secs/related_work.tex
\section{Related Work}
\subsection{End to end Autonomous Driving}
Early end-to-end planners predominantly relied on deterministic modeling, such as MLP-based regression\cite{uniad,vad,sparsedrive,transfuser} and anchor-based classification\cite{vadv2}, imitating a single expert trajectory. GenAD\cite{genad} imposes a GRU-based generator. To align the trajectory diversity, diffusion policies were introduced to recover trajectories from random noise, thereby better aligning with the inherent uncertainty of autonomous driving. Diffusion planners\cite{diffusionplanner,flowplanner} leverage diffusion policies to jointly perform trajectory prediction and planning. GoalFlow\cite{goalflow} adopts goal-conditioned flow matching to generate trajectories, while DiffusionDrive\cite{diffusiondrive} applies truncated diffusion over multiple anchor modes.

\subsection{VLAs for Autonomous Driving}

Conventional end-to-end driving models are often regarded as black boxes. To enhance interpretability, understanding, and reasoning, vision–language models (VLMs)\cite{internvl,qwenvl} have been increasingly incorporated into autonomous driving systems in recent years. Early explorations primarily focused on high-level scene understanding\cite{omnidrive} and reasoning\cite{RAD-Driver,reason2drive,drivecot}, while the demand for action generation gave rise to a large body of Vision–Language–Action (VLA) models.
Early two-stage VLAs\cite{gptdriver,emma,solve,drivevlm} generate meta-actions or low-frequency actions from a VLM, followed by an e2e planner for refinement. However, gradient isolation in such pipelines contradicts the principle of end-to-end learning. In contrast, recent one-stage VLAs directly output the final trajectory. For example, OpenDriveVLA\cite{opendrivevla} autoregressively decodes trajectories, while AutoVLA\cite{autovla} augments the action vocabulary with clustered anchors. These approaches feature token-based decoding and are thus referred to as token-based VLAs.
Other approaches incorporate a planner on top of the VLM to facilitate cross-modal alignment. Orion\cite{orion} employs a GRU-based decoder, while recent works\cite{recogdrive,sgdrive,sparseoccvla} increasingly adopt diffusion policies as planners, achieving stronger performance.

\subsection{Reinforcement Learning for Autonomous Driving}
Imitation learning lacks negative supervision from counterexamples, which leads to poor performance in out-of-distribution (OOD) scenarios. To address this limitation, many studies introduce reinforcement learning to improve the generalization of planners. Early works adopt DPO, while AlphaDrive\cite{alphadrive} is the first to introduce GRPO\cite{deepseekmath}. AdaThinkDrive\cite{adathinkdrive} and AutoVLA\cite{autovla} directly apply GRPO to token-based VLAs, whereas ReCogDrive\cite{recogdrive} designs a diff-GRPO. DiffusionDrivev2\cite{diffusiondrivev2} further proposes intra- and inter-anchor truncated GRPO strategies. More recently, another scoring-based reinforcement learning models\cite{drivor,hydra-mdp,gtrs} performs stage-wise reasoning under direct reward guidance, leading to stronger performance.

%% file: secs/method.tex
\section{Method}
\begin{figure}
    \centering
    \includegraphics[width=1.0\linewidth]{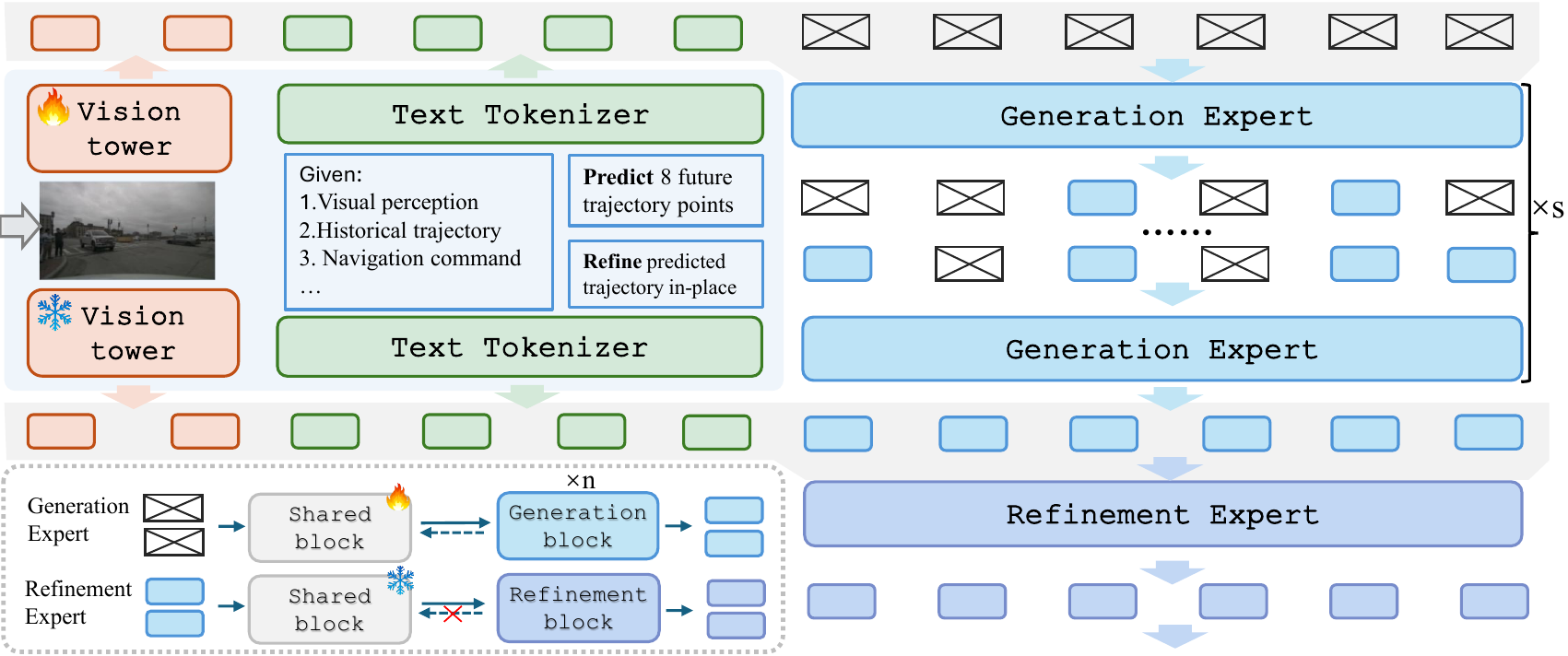}
    \caption{Architecture Overview of DriveFine. Visual and textual inputs are jointly aligned into a unified language space. A set of masked tokens undergoes $s$ steps of parallel denoising followed by a single refinement step. The key difference between the generation expert and the refinement expert lies in their input tokens: the former decodes only the masked tokens, whereas the latter operates on unmasked tokens.}
    \label{fig:overview}
\end{figure}

In this section, we present a detailed introduction of DriveFine with the overview in Fig. \ref{fig:overview}.
Sec. \ref{sec: 3.1} describes the formulation of trajectory planning with the pretrained dLLM.
Sec \ref{sec: 3.2} details the proposed Block-MoE for refinement.
Sec \ref{sec: 3.3} illustrates the hybrid reinforcement learning strategy.

\subsection{Diffusion LLM for Planning}
\label{sec: 3.1}
\subsubsection{Modatily Tokenization.}
As shown in Fig. \ref{fig:overview}, a single front-view image is processed by a pretrained vision tower (SigLIP\cite{siglip}) to produce continuous visual tokens, while a text tokenizer converts textual prompts info discrete tokens. 

To preserve the continuity of trajectory and enable token-level refinement, we discretize the action space following prior works\cite{reflectdrive,wam-diff}.
The spatial range $[-100m,+100m]$ is uniformly divided into 4,000 bins with a resolution of 0.05 m, shared across longitudinal and lateral axes. The heading angle range $[-90°,+90°]$ is discretized into 1800 bins with a resolution of $0.1°$. These bins are appended to the LLM vocabulary, enabling direct decoding of trajectory tokens and facilitating unified cross-modal alignment between language and actions.

\subsubsection{Training and Inference.}
The generating expert follows the standard dLLM training and inference paradigm.
During training, clean sequences are corrupted by random masking, where tokens are replaced by a special mask token [M] with probability $t$, and supervised via masked cross-entropy loss:
\begin{equation}
\mathcal{L}_{\theta}
= - \mathbb{E}_{t, p_0, r_0, r_t}
\left[
\frac{1}{t}
\sum_{i=1}^{L}
\mathbb{I}\!\left[r_t^i = [\text{M}]\right]
\log p_{\theta}\!\left(r_0^i \mid p_0, r_t\right)
\right]
\label{eq:ce}
\end{equation}
Here, $L$, $p_0$, $r_0$ and $r_t$ denote the sequence length, contextual token, original sequence, and the noised sequence, respectively.

During the inference stage, the policy model is encouraged to condition on the given inputs (visions and instructions) and progressively reconstruct a feasible trajectory from fully masked trajectory via several iterative unmasking steps. Specifically, at denoising step $t$, all masked tokens of noised trajectory $r_t$ are predicted in parallel by a mask predictor (generation expert). A subset of mask tokens is then decoded to sequence $r_{t-1}$ for the next iteration. 

\subsection{Block-level MoE for Refinement}
\label{sec: 3.2}
As our previous analysis indicates, the flexible token decoding exacerbates the hazards of irreversible decoding, highlighting the necessity of refinement.


The most straightforward approach is to introduce additional MoE layers to enable adaptive learning, and apply a loss directly on the unmasked tokens for supervision. However, this departs from the standard pre-training and inference paradigm of dLLMs, lossing their foundational capabilities, as they learn only to decode the mask tokens. Moreover, the deep coupling between generation and refinement introduces mutual interference and hinders task-specific tuning and optimization. Yet, fully decoupling the components would inevitably result in a dramatic increase in parameters.



However, despite their different objectives, generation and refinement share a high similarity in contextual representation. Motivated by this insight, we propose our block-level Mixture-of-Experts (block-MoE) together with a carefully designed training–inference pipeline.

As illustrated in Fig. \ref{fig:overview}, a diffusion language model (LLaDA) consists of several stacked blocks.
We treat the pretrained dLLM as a generation expert in its entirety, and just replicate its last
$n$ blocks as refinement blocks, while the preceding blocks and visual tower are shared between both generation and refinement experts.

\subsubsection{Training and Inference.}

During inference, the model is explicitly prompted to perform specific task.
The shared blocks extract common contextual representations, following which the corresponding expert blocks are manually activated to execute either generation or refinement. 

During training, the generating branch computes the loss only on masked tokens as in Eq. \ref{eq:ce}, while the refinement branch computes the loss over all tokens, with gradient flow confined to the refinement expert. Here, the refinement expert undergoes simply warm-started for basic decoding.

Clearly, our block-MoE achieves complete decoupling between generation and refinement during training and inference. This preserves the foundational knowledge of the pretrained model, preventing catastrophic forgetting. Moreover, the refinement expert is plug-and-play and can be trained synchronously with the generation experts, emphasizing its flexibility and transferability.

\subsection{Reinforcement Finetuning}
\label{sec: 3.3}
Recent studies have demonstrated the critical role of reinforcement learning in autonomous driving both theoretically and empirically. In the following, we detail how reinforcement fine-tuning (RFT) is leveraged to explore and enhance the full potential of DriveFine.
\begin{figure}
    \centering
    \includegraphics[width=1.0\linewidth]{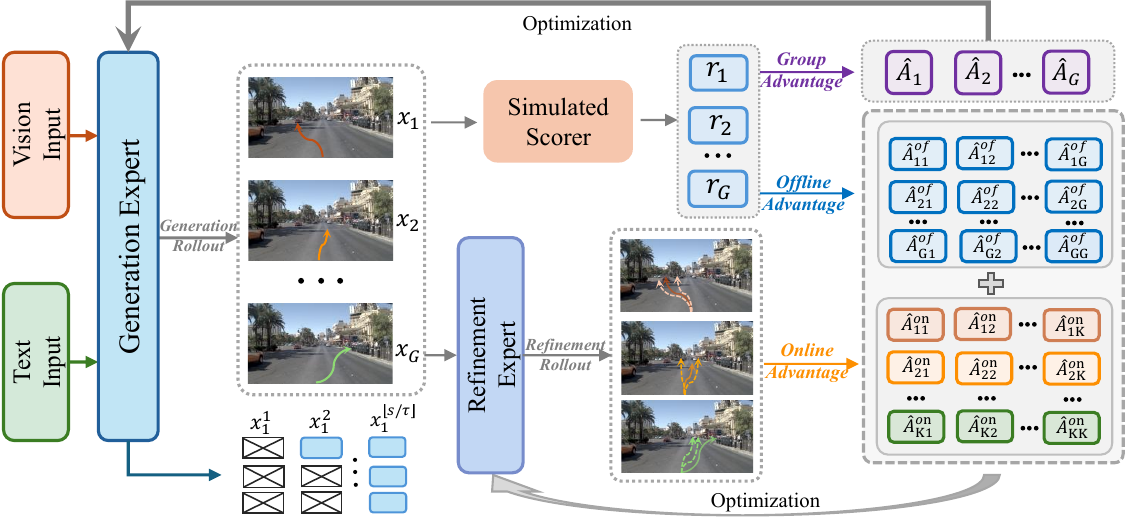}
    \caption{Reinforcement Fine-tuning Pipeline. The offline advantage and online advantage are jointly combined to form the hybrid advantage to supervise the training of the refinement expert.}
    \label{fig:placeholder}
\end{figure}
\subsubsection{GRPO for Generation Expert.}
For the generation expert, we employ a rule-based online reinforcement learning strategy, Group Relative Policy Optimization (GRPO). Specifically, given an arbitrary scenario $p$, the generation expert parallelly samples a group of $G$ candidate trajectories $\{x_i\}_{i=1}^G$. Following \cite{dllm-rl}, we progressively sample for $s$ steps to ensure consistency between training and inference, and aggregate every $\tau$ neighboring steps to balance alignment and efficiency. For each trajectory $x_i$, we retain $\left\lfloor \frac{s}{\tau} \right\rfloor
$ sampled paths $\{x_i^j\}_{j=1}^{\left\lfloor s/\tau \right\rfloor}$. The associated rewards $\{r_i\}_{i=1}^G$ are then evaluated in a simulator. Subsequently, without normalization as in \cite{d1}, the group-relative rewards are computed as follows: 
\begin{equation}
    \hat{A}_i=r_i-\text{mean}(\{r_i\}_{i=1}^G)
\end{equation}
$x_1^{\left\lfloor s/\tau \right\rfloor}$
Subsequently, the sampled tokens are processed sequentially, and their losses are computed using the standard GRPO objective:
\begin{equation}
\small
\begin{split}
\mathcal{L}_{\mathrm{GRPO}}(\theta)
&=
\mathbb{E}_{\substack{q \sim \mathcal{D} \\
o_i\sim \pi_\theta(\cdot \mid q)}}
\Bigg[
\frac{1}{G}
\sum_{i=1}^{G}
\frac{1}{|o_i|}
\sum_{k=1}^{|o_i|}
\min \Big(
\rho_{i,k}^j\,\hat{A}_i^{k},\;
\mathrm{clip}\big(\rho_{i,k}^j,\, 1-\epsilon,\, 1+\epsilon\big)\,\hat{A}_i^{k}
\Big) \\
&\quad
- \beta\, D_{\mathrm{KL}}
\Big(
\pi_\theta(\cdot \mid q)
\;\|\;
\pi_{\mathrm{ref}}(\cdot \mid q)
\Big)
\Bigg],
\end{split}
\end{equation}
where $o_i$ denotes the length of the sequence, $\epsilon$ controls the clipping range, $\beta$ balances the KL divergence penalty, and:

\begin{equation}
\rho_{i,k}^j
=
\left.
\frac{\pi_{\theta}\!\left(o_{i,k}^{j}\mid q,x_i^{j}\right)}
{\pi_{\theta_{\mathrm{old}}}\!\left(o_{i,k}^{j}\mid q,x_i^{j}\right)}
\;\right|\;
x_{i,k}^{j} = [\mathrm{M}],
\; x_{i,k}^{j+1} \neq [\mathrm{M}].
\end{equation}

\subsubsection{Hybrid RL for Refining Expert.}
The purpose of the optimization expert is to fine-tune the generated trajectories to improve their quality. Compared to the anchor trajectories, corrective actions that lead to score improvements should be encouraged, while those that degrade the score should be penalized, regardless of the anchor itself. Therefore, the generated trajectories naturally serve as a reference for advantage computation, eliminating the need for traditional baseline estimation in reinforcement learning (e.g., the value network in PPO or the group-average reward in GRPO). Notably, the sampling of trajectories ensures diversity, allowing them to be directly used for training the optimization expert without further modification. Specifically,for sampled trajectories $\{x_i\}_{i=1}^G$, we compute pairwise reward differences to obtain a relative advantage matrix. Since the sampled trajectories are produced by the generation expert, they constitute offline data for the refining expert; hence, we define this as an offline reward matrix $\hat{\textbf{A}}^\text{of}\in \mathbb{R}^{G\times G}$:

\begin{equation}
    \hat{A}^{\text{of}}_{ij} = r_i - r_j, \quad \forall i,j \in G
\end{equation}

The offline advantage matrix offers several benefits: (1) it has zero mean, simultaneously encouraging improvements and penalizing degradations; (2) the squared advantages provide denser reward signals compared to GRPO, enhancing training stability; and (3) it requires no additional sampling, making the computation simple and efficient.

Despite its many advantages, the optimizer is inherently limited by an upper bound. To encourage the optimizer to explore autonomously, for each generated trajectory $x_i$, we allow it to sample a number of $K$ refined trajectories $\{\hat{x}_{ik}\}_{k=1}^K$ online and compute their corresponding rewards $\{\hat{r}_{ik}\}_{k=1}^K$ simultaneously. The online reward matrix $\hat{\mathbf{A}}^{on}\in\mathbb{R}^{G\times K}$ is then calculated as follows:
\begin{equation}
    \hat{A}^{\text{on}}_{ik}=\hat{r}_{ik}-r_i, \quad \forall i\in G,k\in K
\end{equation}

Finally, a hybrid loss is computed to optimize the refining expert:
\begin{equation}
\small
\begin{split}
\mathcal{L}_{\mathrm{hybrid}}(\theta)
=
&
\mathbb{E}_{\substack{q \sim \mathcal{D} \\
o_i\sim \pi_\theta(\cdot \mid q)}}
\frac{1}{G}
\sum_{i=1}^{G}
\Bigg[
\frac{1}{G}
\sum_{j=1}^{G}
\frac{1}{|o_{ij}|}
\sum_{k=1}^{|o_{ij}|}
\
\rho_{i,k}\,\hat{A}_{of}^{i,j}\;
\\
&+\frac{1}{N}
\sum_{k=1}^{N}
\frac{1}{|o_{ik}|}
\sum_{k=1}^{|o_i|}
\
\rho_{i,k}\,\hat{A}_{on}^{i,j},\;
\Bigg],
\end{split}
\end{equation}

In our implementation, the generator and refiner are trained synchronously: trajectories are sampled online by the generator and then fed to the refiner for training, thereby enabling their collaborative learning and improvement.

%% file: secs/experiments.tex
\section{Experiments}
\subsection{Experimental Settings}
\begin{table*}[t]
\centering
\setlength{\tabcolsep}{4pt}
\renewcommand{\arraystretch}{1.05}
\caption{PDMS results on the \textbf{NAVSIM-v1} benchmark. $^\dagger$ denotes best-of-6 sampling, and * indicates the appliance of score-based reinforcement training as \cite{drivor,gtrs}.}
\begin{tabular}{l|c|ccccc|c}
\toprule
Method & Sensors & NC$\uparrow$ & DAC$\uparrow$ & TTC$\uparrow$ & C.$\uparrow$ & EP$\uparrow$ & PDMS$\uparrow$ \\
\midrule
Human & -- & 100 & 100 & 100 & 99.9 & 87.5 & 94.8 \\
\hline
\rowcolor{gray!15} \multicolumn{8}{c}{\textit{End-to-End Methods}} \\
UniAD\cite{uniad} & 6xC & 97.8 & 91.9 & 92.9 & \textbf{100.0} & 78.8 & 83.4 \\
TransFuser\cite{transfuser} & 3xC+L & 97.7 & 92.8 & 92.8 & \textbf{100.0} & 79.2 & 84.0 \\
LAW\cite{law} & 1xC & 96.4 & 95.4 & 88.7 & 99.9 & 81.7 & 84.6 \\
Hydra-MDP\cite{hydra-mdp} & 3xC+L & 98.3 & 96.0 & 94.6 & \textbf{100.0} & 78.7 & 86.5 \\
DiffusionDrive\cite{diffusiondrive} & 3xC+L & 98.2 & 96.2 & 94.7 & \textbf{100.0} & 82.2 & 88.1 \\
WoTE\cite{wote} & 3xC+L & 98.5 & 96.8 & 94.4 & 99.9 & 81.9 & 88.3 \\
\hline
\rowcolor{gray!15}  \multicolumn{8}{c}{\textit{Vision Language Action Methods}} \\
AutoVLA\cite{autovla} & 3xC & 98.4 & 95.6 & \textbf{98.0} & 99.9 & 81.9 & 89.1 \\

DriveVLA-W0\cite{drivevla-w0} & 1xC & 98.7 & \textbf{99.1} & 95.3 & 99.3 & 83.3 & 90.2 \\
AdaThinkDrive\cite{adathinkdrive} &1xC &98.4 & 97.8 & 95.2 &\textbf{100} & 84.4 &90.3 \\
ReCogDrive\cite{recogdrive} & 1xC & 98.2 & 97.8 & 95.2 & 99.8 & 83.5 & 89.6 \\

DriveFine & 1xC & 98.6 & 97.9 & 95.2 & 99.9 & 85.5 & 90.7\\
\rowcolor{blue!=8}DriveFine* & 1xC & \textbf{98.8} & 98.6 & 96.2 & \textbf{100} & \textbf{86.9} & \textbf{91.8}\\
\midrule
AutoVLA$^\dagger$\cite{autovla} & 3xC & 99.1 & 97.1 & 97.1 & \textbf{100.0} & 87.6 & 92.1 \\
DriveVLA-W0$^\dagger$ & 1xC & \textbf{99.3} & 97.4 & 97.0 & 99.9 & 88.3 & 93.0 \\
AdaThinkDrive$^\dagger$\cite{adathinkdrive} &1xC &99.1 & 98.8 & 97.2 &\textbf{100} & 87.9 &93.0 \\
\rowcolor{blue!=8}DriveFine$^\dagger$ & 1xC & \textbf{99.3} & \textbf{99.2} & \textbf{97.9} & \textbf{100} & \textbf{89.1} & \textbf{94.2} \\
\bottomrule
\end{tabular}
\label{tab:pdms_navtest}
\end{table*}

\subsubsection{Dataset.}
\label{sec: navsim}
We evaluate DriveFine on the NAVSIM\cite{navsim} dataset, which is built upon nuPlan\cite{nuplan} (a subset of OpenScene\cite{openscene}) and provides surround-view images from 8 cameras along with high-quality LiDAR point clouds. The dataset is split into 1,192 (navtrain) scenes for training and 136 scenes (navtest) for testing. NAVSIM also offers a simulation environment for closed-loop evaluation.

NAVSIM v1 adopts the Predictive Driver Model Score (PDMS) as the evaluation metric, which is a weighted aggregation of multiple driving-related criteria, including collision avoidance, drivable area compliance, progress–risk trade-off, and comfort. Building upon this metric, NAVSIM v2 introduces the extended PDM Score (EPDMS), which incorporates additional weighted factors such as traffic light compliance, lane boundary adherence, driving direction following, and extended comfort.
\subsubsection{Implementation Details.}
We adopt Siglip-384\cite{siglip} as the visual tower, which partitions a single front-view image into 8 patches of size $384\times384$. Pretrained weights from LLaDA-8B\cite{llada} are loaded directly, with the first 28 transformer blocks serving as shared blocks and the last 4 blocks designated as expert ones. DriveFine is trained in two stages. In the first stage, it performs supervised fine-tuning (SFT) using the QA pairs and textualized trajectories provided by ReCogDrive\cite{recogdrive}, without any additional pretraining. In the second stage, it undergoes reinforcement fine-tuning (RFT) in the NAVSIM simulation environment.

During the SFT stage, the model is trained for 12 epochs with a batch size of 64, optimized using AdamW with a learning rate of $4\times10^{-5}$ and cosine learning rate decay.

In the RFT stage, The group size for generating expert rollouts is set to 10, each of trajectory is further optimized online by the optimizer 6 times. The model is trained for 1 epoch with a batch size of 16 and a learning rate of $1\times10^{-6}$.

At inference, DriveFine executes 12 sampling steps followed by a single refining step, adhering to a confidence-prioritized and cosine schedule for decoding.

\begin{table*}[t]
\centering
\footnotesize
\setlength{\tabcolsep}{1.45pt}
\caption{EPDMS results on the \texttt{NAVSIM-v2} benchmark. $^\dagger$ indicates results evaluated on the bug-fixed version of NAVSIM.}
\renewcommand{\arraystretch}{1.05}
\begin{tabular}{l|ccccccccc|c}
\toprule
Method & NC$\uparrow$ & DAC$\uparrow$ & DDC$\uparrow$ & TLC$\uparrow$ & EP$\uparrow$ & TTC$\uparrow$ & LK$\uparrow$ & C.$\uparrow$ & EC$\uparrow$ & EPDMS$\uparrow$ \\
\midrule
Ego-MLP\cite{egomlp}        & 93.1 & 77.9 & 92.7 & 99.6 & 86.0 & 91.5 & 89.4 & 98.3 & 85.4 & 64.0 \\
TransFuser\cite{transfuser}     & 96.9 & 89.9 & 97.8 & 99.7 & 87.1 & 95.4 & 92.7 & 98.3 & 87.2 & 76.7 \\
DriveSuprim\cite{drivesuprim}    & 97.5 & 96.5 & 99.4 & 99.6 & 88.4 & 96.6 & 95.5 & 98.3 & 77.0 & 83.1 \\
ARTEMIS\cite{artemis}        & 98.3 & 95.1 & 98.6 & 99.8 & 81.5 & 97.4 & 96.5 & 98.3 & --   & 83.1 \\
ReCogDrive\cite{recogdrive}     & 98.3 & 95.2 & \textbf{99.5} & 99.8 & 87.1 & 97.5 & 96.6 & 98.3 & 86.5 & 83.6 \\
DiffusionDrive\cite{diffusiondrive} & 98.2 & 95.9 & 99.4 & 99.8 & 87.5 & 97.3 & 96.8 & 98.3 & \textbf{87.7} & 84.5 \\
DriveVLA-W0\cite{drivevla-w0} &\textbf{99.0} & \textbf{98.4} &99.3 &99.9 &87.0 &\textbf{98.1} &93.2 &97.9 &58.9 &86.5 \\

\midrule
\rowcolor{blue!8} DriveFine & \textbf{98.7} & 97.3 & 98.8 & 99.8 & 88.2 & 97.8 & \textbf{97.7} & \textbf{98.4} & 84.7 & 87.1 \\

\rowcolor{blue!8} DriveFine$^\dagger$ & \textbf{98.7} & 97.3 & \textbf{99.5} & 99.8 & \textbf{88.7} & 97.8 & \textbf{97.7} & \textbf{98.4} & 83.8 & \textbf{89.7} \\
\bottomrule
\end{tabular}
\label{tab:epdms_navtest}
\end{table*}

\begin{table}[t]
\centering
\fontsize{8.2pt}{10pt}\selectfont
\setlength{\tabcolsep}{0.95pt}
\renewcommand{\arraystretch}{1}
\caption{EPDMS results on the Navhard benchmark. * indicates values are copied from \cite{simscale}; all other results are obtained from our own evaluations.}
\begin{tabular}{p{2.5cm} c ccccccccc | ccccccccc >{\columncolor{gray!20}}c}
\toprule
Method & 
Stage &
 NC$\uparrow$ & DAC$\uparrow$ & DDC$\uparrow$ & TLC$\uparrow$ & EP$\uparrow$ & TTC$\uparrow$ & LK$\uparrow$ & HC$\uparrow$ & EC$\uparrow$
   & EPDMS$\uparrow$\\
\midrule
\multirow{2}{*}{ReCogDrive~\cite{recogdrive}}
& S1
& 97.1 & 80.2 & 98.4 & \textbf{100} & 83.6 & 95.3 & 93.6 & 97.6 & 76.0& 68.9 \\
& S2
& 78.2 & 69.3 & 83.9 & 98.2 & 86.6 & 73.9 & 44.1 & 96.4 & 72.0
& 37.8 \\

\midrule
\multirow{2}{*}{DiffusionDrive*\cite{diffusiondrive}}
& S1
& 96.8 & 86.0 & 98.8 & 99.3 & 84.0 & 95.8 & 96.7 & 97.6 & 79.6 & 66.7 \\
& S2 & 80.1 & \textbf{72.8} & 84.4 & \textbf{98.4} & 85.9 & \textbf{76.6} & 46.4 & 96.3 & \textbf{72.8}
& 40.5 \\
\midrule
\multirow{2}{*}{DriveFine(Ours)} & \cellcolor{blue!8}S1 & \cellcolor{blue!8}\textbf{97.6} & \cellcolor{blue!8}\textbf{90.0} & \cellcolor{blue!8}\textbf{99.1} & \cellcolor{blue!8}99.3 & \cellcolor{blue!8}\textbf{84.9} & \cellcolor{blue!8}\textbf{96.7} & \cellcolor{blue!8}\textbf{97.3} & \cellcolor{blue!8}\textbf{97.6} & \cellcolor{blue!8}72.0 & \cellcolor{blue!8}\textbf{74.4} \\ 

& \cellcolor{blue!8}S2 & \cellcolor{blue!8}\textbf{82.1} & \cellcolor{blue!8}71.3 & \cellcolor{blue!8}\textbf{84.8} & \cellcolor{blue!8}\textbf{98.4} & \cellcolor{blue!8}\textbf{88.1} & \cellcolor{blue!8}74.3 & \cellcolor{blue!8}\textbf{47.2} & \cellcolor{blue!8}\textbf{96.8} & \cellcolor{blue!8}\textbf{72.8} & \cellcolor{blue!8}\textbf{41.0} \\
\bottomrule
\end{tabular}
\label{tab:navhard}
\end{table}

\subsection{Main Results}
\subsubsection{Results of Navtest.}
We first report the NAVSIM v1 results, evaluating PDMS on the navtest split, as shown in Tab. \ref{tab:pdms_navtest}. DriveFine achieves state-of-the-art (SOTA) performance. In particular, compared with autoregressive token-based VLA models\cite{autovla,adathinkdrive}, DriveFine achieves a 0.5\% improvement, while being on par with diffusion-based planners\cite{recogdrive}. When score-based reinforcement fine-tuning is applied (an additional scorer is trained), DriveFine further improves its performance to 91.9 PDMS, surpassing all existing VLA planners.

We report the comparison results on NAVSIM-v2 in Tab. \ref{tab:epdms_navtest}, which provides a more comprehensive evaluation of the overall model performance. For a fair comparison with prior works, we evaluate DriveFine using an earlier NAVSIM version, which contains a score computation bug that leads to systematically underestimated overall scores. Despite this issue, DriveFine still achieves 86.7 EPDMS, outperforming DriveVLA-W0\cite{drivevla-w0} by 0.5 points.
When evaluated with the bug-fixed NAVSIM version, DriveFine further reaches 89.7 EPDMS, thereby establishing a new state of the art.

\subsubsection{Results of Navhard.}
We further conduct experiments on the more challenging Navhard benchmark, which employs Gaussian splatting to generate scenarios beyond the training data distribution and adopts a two-stage evaluation protocol. The results are reported in Tab. \ref{tab:navhard}. Notably, no additional training is applied in this setting. Under this fair evaluation protocol, DriveFine outperforms the diffusion-based methods\cite{diffusiondrive,reason2drive} on both stage metrics, with a particularly notable improvement of 5.5 points in Stage 1 EPDMS over ReCogDrive, further demonstrating its strong performance as well as the superior generalization capability of token-based VLAs.

\subsection{Ablation Studies}
\subsubsection{Ablations of the core components.}
Tab. \ref{tab:ablation} presents an ablation study on the core components of DriveFine. With LLaDA mimicking expert trajectories (SFT), the model achieves a PDMS of 86.7. GRPO reinforcement training increases PDMS to 89.6. Incorporating the refinement expert with offline reinforcement training further improves performance by 0.7 point, while online reinforcement training raises the model's performance ceiling to 90.8. 

\begin{figure}
    \centering
    \includegraphics[width=1.0\linewidth]{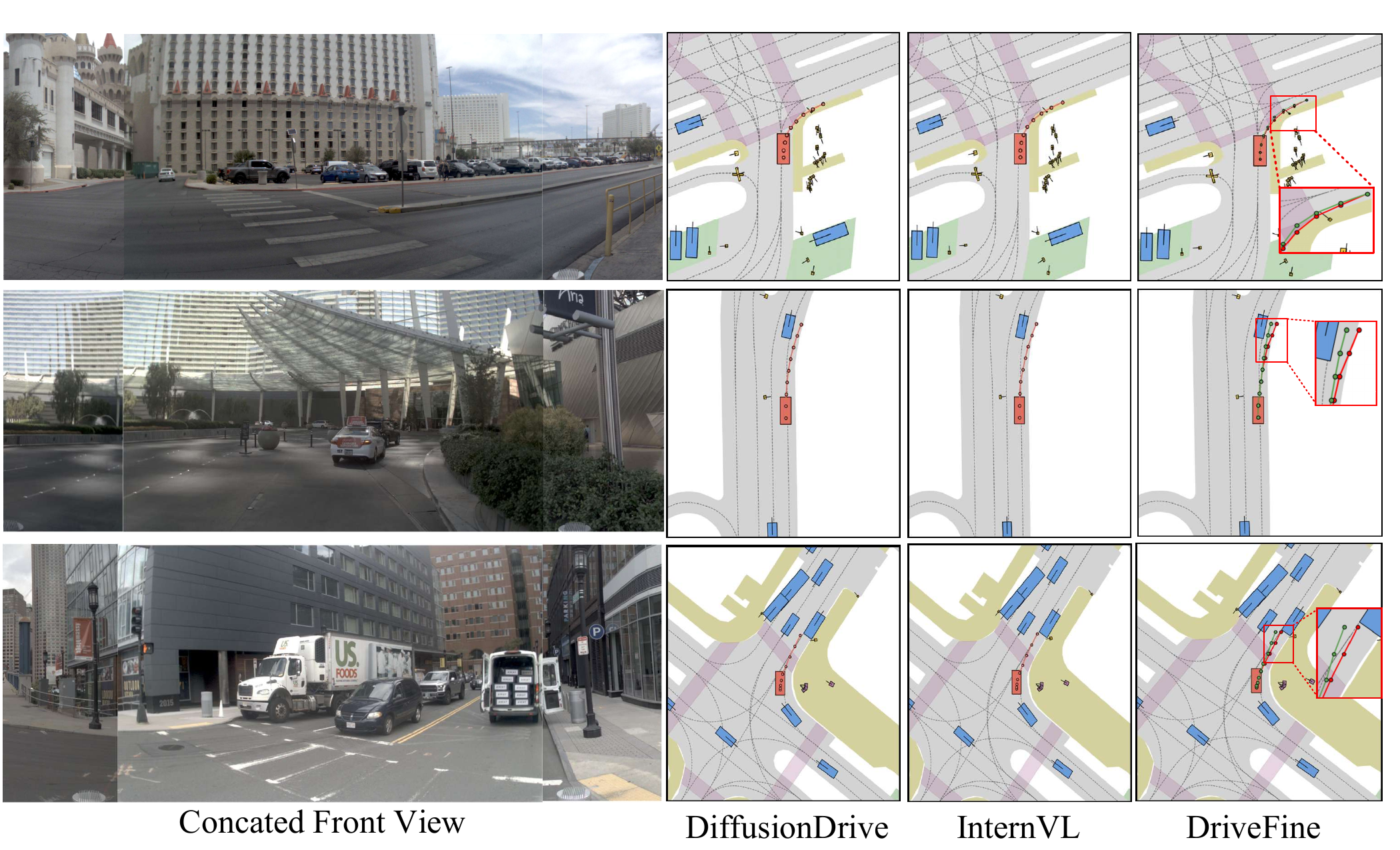}
    \caption{Qualitative visualization comparison between DriveFine and other SOTA methods. In the rightmost figure, the {\color[HTML]{FF0000}red} and {\color[HTML]{59A14F}green} lines denote the trajectories before and after refinement, respectively.}
    \label{fig:vis}
\end{figure}

We observe that the refinement mechanism improves nearly all metrics, particularly DAC and Conf. Further visualization of trajectories before and after refinement (as shown in the Fig. \ref{fig:vis}) clearly demonstrates that the refinement expert can effectively correct anomalies when individual tokens lead to collisions or off-road events, which would otherwise result in complete trajectory failure. Additionally, it significantly mitigates fluctuations caused by noncausal decoding, enhancing overall trajectory smoothness.

\begin{table}[t]
\centering
\setlength{\tabcolsep}{3pt}
\renewcommand{\arraystretch}{1.0}
\caption{Ablation results of core components of DriveFine.}
\begin{tabular}{c c c c c c c c c c >{\columncolor{blue!10}}c}
\toprule
ID &
\begin{tabular}[c]{@{}c@{}}SFT\end{tabular} &
\begin{tabular}[c]{@{}c@{}}GRPO\end{tabular} &
\begin{tabular}[c]{@{}c@{}}Offline-RFT\end{tabular} &
\begin{tabular}[c]{@{}c@{}}Online-RFT\end{tabular} &
NC & DAC & TTC & Conf. & EP &
\textbf{PDMS} \\
\midrule
1 & \cmark & \xmark & \xmark & \xmark & 98.0 & 95.2 & 94.9 & 99.4 & 81.6 & 86.7 \\
2 & \cmark & \cmark & \xmark & \xmark & 98.3 & 96.9 & 95.1 & 99.2 & 85.1 & 89.6\textcolor{blue}{{\tiny +2.9}} \\
3 & \cmark & \cmark & \cmark & \xmark & 98.5 & 97.3 & 95.2 & 99.9 & 85.4 & 90.3\textcolor{blue}{{\tiny +0.7}} \\
4 & \cmark & \cmark & \cmark & \cmark & 98.6 & 97.9 & 95.2 & 99.9 & 85.5 & 90.8\textcolor{blue}{{\tiny +0.5}} \\
\bottomrule
\end{tabular}

\label{tab:ablation}
\end{table}

\subsubsection{Ablation study on the robustness of DriveFine.}
We further evaluated the robustness of DriveFine. As shown in Tab. \ref{tab:robustness}, after PDMS-oriented RFT, extended metrics such as LK, EC, and DDC remain stable or show slight improvements. Consequently, the increase in PDMS is accompanied by a synchronous improvement in EPMDS. Compared with diffusion-based models in \ref{fig:pdms}, DriveFine demonstrates a clear robustness advantage, further confirming the potential of token-based models.


\begin{table}[t]

\begin{minipage}[t]{0.52\linewidth}
\centering
\captionof{table}{Performance comparison of different models.}
\fontsize{7.5}{12}\selectfont 
\renewcommand{\arraystretch}{0.9}
\begin{tabular}{l | c c c >{\columncolor{blue!8}}c >{\columncolor{blue!8}}c}
\toprule
Model & EP$\uparrow$ & LK$\uparrow$ & EC$\uparrow$ & PDMS$\uparrow$ & EPMDS$\uparrow$ \\
\midrule
DriveFine-SFT & 86.8 & 97.9 & 83.3 & 86.7 & 86.8 \\
+RFT(PDMS) & 88.7 & 97.7 & 83.8 & 90.5\textcolor{blue}{{\tiny +3.8}} & 89.1\textcolor{blue}{{\tiny +2.3}} \\
\bottomrule
\end{tabular}
\label{tab:robustness}
\end{minipage}
\hfill
\begin{minipage}[t]{0.48\linewidth}
\centering
\caption{Sensitivity the number of refinement blocks $n$.}
\label{tab:efficiency}
\setlength{\tabcolsep}{2pt}
\renewcommand{\arraystretch}{1}
\begin{tabular}{c | c c c c c c}
\toprule
n & 0 & 1 & 2 & \cellcolor{blue!8}4 & 6 & 8 \\
\midrule
Param(B) & 8 & 8.25 & 8.5 & \cellcolor{blue!8}9 & 9.5 &10 \\
PDMS & 89.6 & 90.0 & 90.4 & \cellcolor{blue!8}90.8 & 90.8 &90.7 \\
\bottomrule
\end{tabular}
\label{tab:n}
\end{minipage}

\end{table}

\subsubsection{Sensitivity analysis of the number of refinement blocks.}
We evaluated the sensitivity of DriveFine to the number of refinement blocks, as summarized in Tab. \ref{tab:n}. When $n=0$, i.e., without any refinement capability, the trajectories are derived from the generative expert. Remarkably, introducing just one refinement block (250M parameters) already improves trajectory quality (+0.4 PDMS). As the number of blocks increases, performance gradually improves, reaching the optimal level when 4 refinement blocks are injected.

\subsubsection{Efficiency-Performance Trade-off Analysis.}

\begin{wrapfigure}{r}{0.45\textwidth}
    \vspace{-10pt} 
    \centering
    \includegraphics[width=\linewidth]{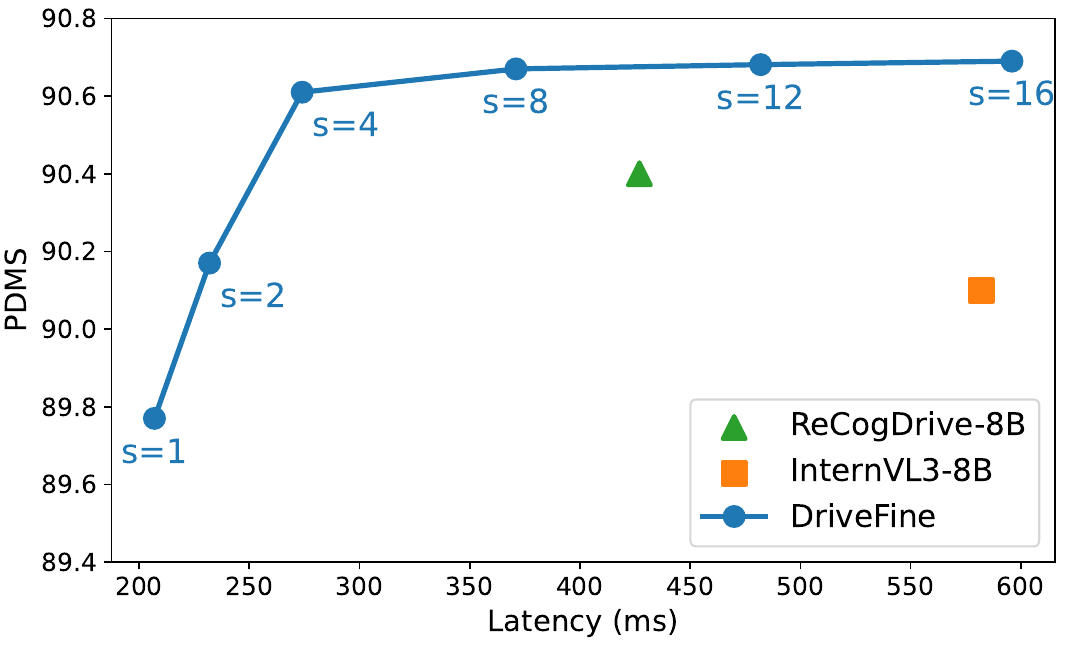}
    \caption{PDMS-Latency trade-off}
    \label{fig:pdms}
    \vspace{-10pt}
\end{wrapfigure}

We evaluate the impact of the number of diffusion steps $s$ on the efficiency–performance trade-off, comparing DriveFine with VLA models of similar scale, as shown in Fig. \ref{fig:pdms}.
As expected, inference latency increases roughly proportionally with $s$. In general, a larger number of diffusion steps allows the model to “think” more thoroughly, resulting in better performance. 

However, we observe that with only 4 steps, DriveFine already achieves a PDMS of 90.47, comparable to ReCogDrive-8B\cite{recogdrive}, with an average latency of merely 207 ms, representing an optimal efficiency–performance balance. This further demonstrates the robustness of masked diffusion LLMs.
In contrast, the InternVL model relies on token-by-token decoding, incurring much higher inference latency.
Overall, these results indicate that the flexible decoding of diffusion-based language models enables an effective balance between efficiency and performance, paving the way for more efficient planning paradigms.

\subsubsection{Effect of Group Size.}
As shown in Tab. \ref{tab:group_size}, DriveFine is moderately sensitive to the GRPO group size. Even with a small group size $G=2$, it outperforms the SFT baseline by 1.7 PDMS. Increasing $G$ leads to a monotonic performance gain, with the best performance achieved at $G=8$.

\begin{table}
\centering
\begin{minipage}{0.48\linewidth} 
\centering
\caption{Sensitivity of group size $G$.}
\fontsize{8.8}{12}\selectfont 
\renewcommand{\arraystretch}{1.2}
\begin{tabular}{c|cccccc}
\toprule
Group size $G$ & 0 & 2 & 4 & 6 & \cellcolor{blue!8}8 & 10\\
\midrule
EP$\uparrow$    &  81.6  &  83.7   &  84.4  &  84.7  & \cellcolor{blue!8}85.1  &85.2\\
DAC$\uparrow$    &  95.2  & 95.8  & 96.3  &  96.5 & \cellcolor{blue!8}96.9 &96.8\\
PDMS$\uparrow$   & 86.7 & 88.4 & 89.1 & 89.4 & \cellcolor{blue!8}89.6 &89.6  \\
\bottomrule
\end{tabular}
\label{tab:group_size}
\end{minipage}
\end{table}